\documentstyle[aclap,epic]{article}         
\begin{document}
\title{Application-driven automatic subgrammar
  extraction}
\author{Renate Henschel\\ Centre for Cognitive Science \\ 
2 Buccleuch Place, Edinburgh, UK \\({\tt henschel@cogsci.ed.ac.uk}) \And
John Bateman \\Language and Communication Research\\University of
Stirling,
Stirling, UK\\ ({\tt j.a.bateman@stir.ac.uk})}

\bibliographystyle{acl}
\maketitle

\vspace*{-3in}{\fbox{\parbox[t]{4in}{\small From: ACL-EACL'97 Workshop
on {\em Computational Environments for Grammar Development and Linguistic
Engineering} (eds: Estival, Lavelli, Netter and Pianesi), Madrid, 1997.
pp46--53.}}}

\vspace*{+2.5in}

\begin{abstract}
{The  space  and   run-time  requirements of    broad coverage
grammars appear for  many applications unreasonably large in  relation
to  the relative simplicity of the  task at hand.   On the other hand,
handcrafted development of application-dependent grammars is in danger
of  duplicating work   which is then  difficult  to   re-use in  other
contexts of application.  To overcome this problem, we present in this
paper a procedure  for the  automatic extraction of  application-tuned
consistent subgrammars from  proved  large-scale  generation grammars.
The procedure has been implemented for large-scale systemic 
grammars  and builds on the formal equivalence between systemic grammars and typed unification based grammars.      Its
evaluation for the generation of encyclopedia entries is described, 
and directions
of future development, applicability, and extensions are discussed.
}\footnote{This work was partially supported by the DAAD through grant D/96/17139.}

\end{abstract}

\section{Introduction}

Although  we have reached a   situation  in computational  linguistics
where  large coverage grammars  are  well developed  and  available in
several formal traditions, the use of these research results in actual
applications and  for   application   to specific domains     is still
unsatisfactory.  One  reason for  this   is that  large-scale  grammar
specifications incur a seemingly unnecessarily  large burden of  space
and processing  time that  often does  not  stand in relation  to  the
simplicity of the particular task.  The usual alternatives for natural
language  generation to date  have been the handcrafted development of
application or  sublanguage specific grammars or   the use of template
based  generation grammars.  In   \cite{Busemann96} 
both approaches are
combined resulting in a practical small generation grammar
tool. But still the grammars are handwritten or, if extracted from
large grammars, must be adapted by hand. In general, both -- the template and the
handwritten application grammar approach -- compromise  the idea of a
general {\sc nlp} system architecture with  reusable bodies of general
linguistic resources.   

We   argue that this customization  bottleneck  can be overcome by the
automatic    extraction  of  application-tuned consistent   generation
subgrammars from proved given  large-scale grammars.  In this paper we
present such an automatic subgrammar  extraction tool.  The underlying
procedure  is   valid   for grammars  written   in  typed  unification
formalisms; it is here carried  out for  systemic grammars within  the
development environment for text generation {\sc kpml} \cite{KPML1}.
The input is a set  of semantic  specifications covering the  intended
application.  This can either be  provided  by generating a predefined
test suite or be   automatically produced  by running the   particular
application during a training phase.

The paper is structured as follows.  First, an algorithm for automatic
subgrammar extraction for  arbitrary systemic grammars will  be given,
and  second the application  of  the algorithm  for generation  in the
domain of `encyclopedia entries'  will be illustrated.  To conclude, we
discuss several  issues raised  by  the work described, including  its
relevance   for typed unification based   grammar descriptions and the
possibilities for further improvements in generation time.

\section{Grammar extraction algorithm}

Systemic Functional   Grammar (SFG)~\cite{Halliday85} is  based on the
assumption that the differentiation  of syntactic phenomena is  always
determined   by  its function  in  the   communicative context.   This
functional orientation has lead to the creation of detailed linguistic
resources that are     characterized by an  integrated   treatment  of
content-related,    textual  and  pragmatic    aspects.  Computational
instances of systemic grammar are successfully employed in some of the
largest and most influential text  generation projects---such as,  for
example, PENMAN~\cite{Mann83}, COMMUNAL~\cite{FawcettTucker90-coling},
TECHDOC~\cite{RoesnerStede94-coling},
Drafter~\cite{ParisVanderLinden96}, and Gist~\cite{NotStock94}.

For our present purposes, however, it is the formal characteristics of
systemic  grammar and  its implementations  that  are more  important.
Systemic   grammar      assumes          multifunctional   constituent
structures  representable  as feature 
structures  with coreferences. As shown in the following function
structure example for the sentence ``The people that buy silver love it.'', different functions can be filled by one and the
same constituent:

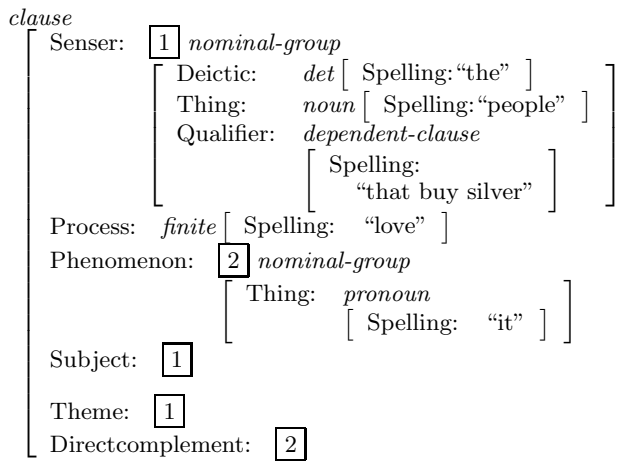
\vspace{0.15in}
\begin{footnotesize}
\hspace*{-0.2in}\mbox{\footnotesize \em clause}\\[-0.15in]
\[ \left[ \begin{array}{ll}
\mbox{Senser:} \ \ \  \fbox{\footnotesize 1}~\mbox{\footnotesize \em
  nominal-group} \\
\hspace*{0.5in} \left[ \begin{array}{ll}
\mbox{Deictic:} & \mbox{\footnotesize \em det}
 \left[ \begin{array}{l}
\mbox{Spelling:} \mbox{``the''} \end{array} \right] \\
\mbox{Thing:} & \mbox{\footnotesize \em noun}
 \left[ \begin{array}{l}
\mbox{Spelling:} \mbox{``people''} \end{array} \right] \\
\mbox{Qualifier:} & \mbox{\footnotesize \em dependent-clause} \\
 &  \left[ \begin{array}{l}
\mbox{Spelling:} \\
  \mbox{\ \ \ ``that buy silver''} \end{array} \right]
\end{array} \right]
 \\
\mbox{Process:}\ \ \    \mbox{\footnotesize \em finite}
 \left[ \begin{array}{ll}
\mbox{Spelling:} & \mbox{``love''} \end{array} \right] \\
\mbox{Phenomenon:} \ \ \ \parbox[t]{2in}{$ \fbox{\footnotesize 2}~\mbox{\footnotesize \em
  nominal-group} \\
  \left[ \begin{array}{ll}
\mbox{Thing:} & \mbox{\footnotesize \em pronoun} \\
              & \left[ \begin{array}{ll}
\mbox{Spelling:} & \mbox{``it''} \end{array} \right] \end{array}
\right] $ }  \\
\mbox{Subject:} \ \ \   \fbox{\footnotesize 1} \\[0.1in]
\mbox{Theme:} \ \ \  \fbox{\footnotesize 1} \\
\mbox{Directcomplement:}\ \ \  \fbox{\footnotesize 2} \\
\end{array} \right] \]

\end{footnotesize}

Given  the  notational equivalence of
HPSG and  systemic grammar first mentioned  by \cite{Carpenter92} 
and \cite{Zajac92-cl}, 
and further elaborated
in \cite{Henschel95-edin}, one can characterize  a systemic grammar as
a large type hierarchy with multiple (conjunctive and disjunctive) and
multi-dimensional inheritance with an open-world semantics.  The basic
element of a   systemic  grammar---a  so-called {\em  system}---is   a
type axiom    of  the form (adopting   the  notation  of {\sc
cuf}~\cite{CUF96}):

\begin{verbatim}
 entry = type_1 | type_2 | ... | type_n.
\end{verbatim}

\noindent where $type_1$ to $type_n$  are  exhaustive and disjoint  subtypes  of
type $entry$. $entry$ need not necessarily be a single type; it can be
a logical expression over types  formed with the connectors {\sc  and}
and {\sc  or}.  A  systemic grammar therefore  resembles  more  a type
lattice than    a type hierarchy  in the   HPSG tradition. In systemic
grammar, these basic type axioms,  the systems, are named; we
will use $entry(s)$ to denote the left-hand  side of some named system
$s$, and $out(s)$ to denote   the set of subtypes \{$type_1,   type_2,
..., type_n$\}-- the output of the system. The following type axioms taken from the large
systemic English grammar {\sc nigel} \cite{Matthiessen83-eacl} shall illustrate
the nature of systems in a systemic grammar:
 
{\small 
\begin{verbatim}
nominal_group = class_name | individual_name.
nominal_group = wh_nominal | nonwh_nominal.
(OR class_name wh_nominal) = singular | plural.
\end{verbatim}}

\noindent The meaning of these type axioms is fairly obvious: 
  Nominal groups can be subcategorized in class-names and 
  individual-names on the one hand, they can be subcategorized with
  respect to their WH-containment into WH-containing nominal-groups
  and nominal-groups without WH-element on the other hand. The singular/plural
  opposition is valid for class-names as well as for WH-containing nominal
  groups (be they class or individual names), but not for
  individual-names without WH-element. 

  Systemic   types  inherit  constraints with    respect to
appropriate  features,  their filler   types, coreferences and  order.
Here are the constraints for some of the types defined above:

\begin{quote}
{\small 
{\em nominal-group} [Thing: {\em noun}]\\
{\em class-name} [Thing: {\em common-noun},\\
 \hspace*{0.7in}Deictic: {\em top}]\\
{\em individual-name} [Thing: {\em proper-noun}]\\
{\em wh-nominal} [Wh: {\em top}] }
\end{quote}

Universal principles  and rules  are  in systemic grammar not factored out.    The lexicon
contains stem forms  and has a  detailed word class type  hierarchy at
its top.  Morphology is also organized  as a monotonic type hierarchy.
Currently used  implementations      of SFG are   the   PENMAN  system
\cite{PENMAN89},         the        KPML   system~\cite{KPML1}     and
WAG-KRL~\cite{ODonnell94}.  

Our subgrammar extraction has been applied and tested in the context of the
KPML  environment.  KPML adopts the processing  strategy of the PENMAN
system  and so it  is necessary   to  briefly describe  this strategy.
PENMAN    performs a semantic   driven  top-down traversal through the
grammatical  type hierarchy for every  constituent.   Passed types are
collected and their   feature   constraints are unified  to    build a
resulting   feature  structure.   Substructure  generation requires an
additional grammar traversal controlled by the feature values given in
the superstructure. In addition to the grammar  in its original sense,
the PENMAN system provides a particular  interface between grammar and
semantics. This interface is organized with the help of so-called {\em
choosers}---these are decision  trees associated with  each system 
 of the grammar  which  control the  selection of an   appropriate
subtype  during traversal.  Choosers should    be seen as a  practical
means of enabling  applications (including text  planners) to interact
with  the grammar using purely  semantic  specifications even though a
fully  specified semantic theory may  not yet be available for certain
important areas necessary for coherent,  fluent text generation.  They
also serve to enforce deterministic choice---an important property for
practical generation (cf.~\cite{Reiter94}).

The basic form of a chooser node is as follows.
\begin{quote}
({\bf ask} {\em query}\\
\hspace*{0.6 in}     ({\em answer1 actions})\\
 \hspace*{0.6in}    ({\em answer2 actions})\\
\hspace*{0.6in}     ...)
\end{quote}
The nodes in  a chooser  are  queries to  the semantics, the  branches
contain a set of actions including  embedded queries. Possible chooser
actions are the following:
\begin{quote}
({\bf ask} {\em query} (..) ... (..))\\
({\bf choose} {\em type})\\
({\bf identify} {\em function concept})\\
({\bf copyhub} {\em function1 function2})
\end{quote}

\noindent  A choose action of a  chooser explicitly ({\bf choose} {\em
 type}) selects one of the output types of  its associated system.  In
 general, there can be several paths through a given chooser that lead
 to the  selection  of  a single   grammatical type: each   such  path
 corresponds  to a  particular  configuration  of semantic  properties
 sufficient to motivate the  grammatical type selected.  Besides this,
 choosers serve to create a binding between given semantic objects and
 grammatical constituents to  be generated.  This  is performed by the
 action ({\bf identify} {\em function} {\em concept}).  Because of the
 multifunctionality assumed for the  constituent structure in systemic
 grammar, two  grammatical functions can   be realized by one and  the
 same constituent  with one  and the same  underlying  semantics.  The
 action ({\bf copyhub}  {\em function1 function2}) is  responsible for
 identifying the semantics of both grammatical functions.

Within such a framework, the  first stage of subgrammar extraction  is
to ascertain  a representative set of   grammatical types covering the
texts for the  intended application.  This can  be obtained by running
the text  generation  system  within  the  application with   the full
unconstrained  grammar.    All grammatical   types   used during  this
training stage are collected  to form the  backbone for the subgrammar
to   be  extracted.   We call   this   cumulative  type set the   {\em
goal-types}.

\begin{figure*}[t]
\fbox{\parbox[t]{6in}{\noindent extract-subgrammar($goaltypes$)

\noindent 1 \hspace*{0.1in} {\bf for all} $s \in systems$ \\
\hspace*{0.3in} {\bf do} $entry(s)$ :=
remove-unsatisfiable-features($entry(s)$)\\
2 \hspace*{0.1in}$*subgrammar* := \emptyset$\\
3 \hspace*{0.1in}traverse-system($rank, start, start, \emptyset, goaltypes$)\\

\noindent traverse-system($s, type, supertype, inheritedconstraints, goaltypes$)

\noindent 1 \hspace*{0.1in} $inter$ := $out(s) \cap goaltypes$\\
2 \hspace*{0.1in} {\bf if} $inter \neq \emptyset$  \\
  \hspace*{0.3in}   {\bf then if} $|entry(s)|$ = 1 {\bf and} $|inter|$ = 1\\
 \hspace*{0.8in}{\bf then} {\bf do} $out$ := the single  element in $inter$\\
\hspace*{1.4in}$constraints$ := unify($constraints(out),inheritedconstraints$)\\
\hspace*{1.4in}traverse-type($out, supertype, constraints, goaltypes$)\\
 \hspace*{0.8in}{\bf else} {\bf do} $entry(s)$ := dnf-substitute($supertype, type, entry(s)$)\\
\hspace*{1.4in}$out(s)$ := $inter$\\
\hspace*{1.4in}push($s, *subgrammar*$)\\
\hspace*{1.4in}{\bf for all} $out \in inter$\\
\hspace*{1.5in}{\bf do} traverse-type($out, out$, $\emptyset$, $goaltypes$)\\
\hspace*{1.7in}$constraints(supertype$) := \\
\hspace*{2in}unify($constraints(supertype$),$inheritedrealizations$)\\

\noindent traverse-type($type, supertype, inheritedconstraints, goaltypes$)\\
1 \hspace*{0.1in}$who$ := who-has-in-entry($type$)\\
2 \hspace*{0.1in}{\bf if} \hspace*{0.1in}$who$ = $\emptyset$ {\bf and} $inheritedconstraints \neq \emptyset$\\
\hspace*{0.3in}{\bf then do} $constraints(supertype$) := \\
\hspace*{0.6in}unify($constraints(supertype$), $inheritedconstraints$)\\
3 \hspace*{0.1in}{\bf for all} $s \in who$\\
\hspace*{0.4in} {\bf  do} traverse-system($s, type, supertype,
inheritedconstraints, goaltypes$)}}
\caption{Subgrammar extraction algorithm}
\label{alg}
\end{figure*}

The list of {\em goal-types} then gives the point of departure for the
second  stage, the automatic extraction   of a consistent  subgrammar.
{\em goal-types} is used as  a filter against which systems (type
axioms)  are  tested. 
Types  not  in {\em goal-types} have to be excised from the subgrammar being
extracted. This is carried out for the entries of the systems in a
preparatory step.  We assume that the    entries are given in   disjunctive
normal form.  First, every conjunction  containing a type which is not
in {\em goal-types} is removed.  After  this deletion of unsatisfiable
conjunctions, every type in an entry which is not  in {\em goal-types} is removed.
The restriction of the outputs of every system to the
 {\em goal-types} is done during   a simulated  depth-first  traversal
through   the entire grammatical   type lattice.
The procedure works on the type lattice with the revised
entries. Starting with the  most  general type  {\em start} (and  the most
general system called {\em rank} which  is the system with {\em start}
as  entry),  a hierarchy traversal  looks  for  systems which although
restricted to  the type set {\em goal-types}  actually  branch,
i.e. have more than one type in their output.  These
systems constitute the  new subgrammar.  In essence, each  grammatical
system $s$ is examined to see how  many of its  possible subtypes in
$out(s)$ are used
within the target  grammar. Those types which are  not used are excised from
the  subgrammar  being   extracted.   More specific     types that are
dependent  on any excised types  are not considered further during the
traversal.  Grammatical systems where there is only a single remaining
unexcised   subtype   collapse to  form    a degenerated pseudo-system
indicating that no grammatical 
variation is possible in the considered application domain. For example, in the
application described in section 3 the system

{\tt \small indicative = declarative $|$ interrogative.}

\noindent collapses into

{\tt \small indicative = declarative.}

\noindent because  questions do not occur in the application
domain. Pseudo-systems 
of this kind are not  kept in the subgrammar. The
 types on their right-hand side (pseudotypes) are excised accordingly, although
 they are used for deeper traversal, thus defining  a path to more
 specific systems.  Such a path can consist of more than one
 pseudotype, if the repeated traversal steps find further degenerated
 systems. Constraints defined for pseudo-types are raised, chooser actions
are percolated down---i.e., more  precisely, constraints  belonging to
 a pseudo-type
 are unified with the constraints of  the most general not pseudo type
 at the beginning of 
the path. Chooser actions  from systems on the path  are collected and extend the
chooser associated  with the final (and first not pseudo) system of the
path.  However, in the case that a maximal type is reached which is not in {\em
goal-types}, chooser actions have to be raised too. The number of {\em
goal-types} is then usually larger than the
number of the types  in the extracted subgrammar  because all pseudotypes in
{\em goal-types}   are
excised.

As the recursion criteria in the traversal, we first simply look for a
system which  has the actual type in  its revised entry regardless of the fact
if it  occurs  in a conjunction or  not.   This on its   own, however,
oversimplifies the  real logical relations between  the types and would
create an   inconsistent subgrammar.  The  problem is  the conjunctive
inheritance. If the current type occurs  in an entry of another system
where it  is  conjunctively bound, a  deeper  traversal  is in  fact only
licensed   if the  other types  of  the   conjunctions are chosen  as
well. In order  to perform such a  traversal, a breadth traversal with
compilation of all crowns of the lattice (see \cite{Ait-Kaci89}) would
be necessary.  In order to avoid this potentially computationally very
expensive operation,   but  not to  give   up the consistency  of  the
subgrammar, the  implemented subgrammar  extraction procedure sketched
in  Figure~\ref{alg}  maintains  all systems   with complex  entries (be they
conjunctive  or disjunctive) for  the subgrammar even   if they do not
really       branch    and       collapse    to   a     single-subtype
system.\footnote{Keeping the  disjunctive systems is not necessary for
the consistency,  but  saves  multiple  raising  of one and   the same
constraint.} A related approach can be found in \cite{ODonnell92-eda}
for the extraction of smaller systemic subgrammars for analysis.

If the lexicon is organized as or under a  complex type hierarchy, the
extraction of an application-tuned lexicon  is carried out  similarly.
This has  the  effect that closed  class words   are  removed from the
lexicon if they are not covered in the application domain.  Open class
words belonging to word classes not covered by the subgrammar type set
are removed.  Some applications do not need their own lexicon for open
class  words because they  can  be linked   to an  externally provided
domain-specific thesaurus  (as     is  the case  for  the   examples
discussed  below).  In  this   case, a  sublexicon  extraction  is not
necessary.

\section{Application for text type `lexicon biographies'}

The first trial  application  of the automatic  subgrammar  extraction
tool has  been carried out for  an  information system  with an output
component that  generates    integrated text    and  graphics.    This
information  system has been developed for   the domain of art history
and is capable of providing short biography articles for around 10~000
artists.  The  underlying  knowledge base,  comprising half  a million
semantic concepts, includes  automatically extracted information  from  14~000
encyclopedia articles from  McMillans planned publication ``Dictionary
of Art'' combined with several  additional information sources such as
the Getty  ``Art and  Architecture  Thesaurus''; the   application  is
described in detail in~\cite{Kamps-etal96}.   As input the user clicks
on an  artist name.  The system  then performs content selection, text
planning, text and diagram generation  and page layout  automatically.
Possible output languages are  English and German.  

The grammar necessary   for short biographical   articles is, however,
naturally much  more    constrained than that  supported   by  general
broad-coverage grammars.  There are  two main reasons for this: first,
because of the relatively  fixed text type  ``encyclopedia biography''
involved, and second,  particularly in the example information system,
because of the relatively  simple nature of the knowledge  base---this
does not support more sophisticated text generation as might appear in
full encyclopedia articles.  Without extensive empirical analysis, one
can already state  that such a grammar  is restricted to main clauses,
only coordinative    complex   clauses,  and  temporal   and   spatial
prepositional phrases.  It  would  probably  be possible  to produce
the generated texts  with relatively complex templates and aggregation
heuristics: but  the full   grammars  for English and German available
in KPML already  covered  the   required
linguistic phenomena.

The  application of the automatic subgrammar  extraction  tool to this
scenario is as follows.

\begin{figure}[t]
\rule{\columnwidth}{0.2mm}
\begin{center}
\begin{picture}(210,140)
\put(0,0){\scriptsize\sffamily \grid(210,140)(30,10)[0,120]}
\thicklines
\put(0,-120){\putfile{bio-stat2.put}{\tiny $\times$}}
\end{picture}
\end{center}

Example texts:

{\small \begin{quote} 
Roger Hilton was  an English painter. He  was born at  Northwood on 23
March 1911, and he  died at Botallack on 23  February 1975. He studied
at Slade  School in 1929  - 1931. He  created "February - March 1954",
"Grey figure", "Oi yoi yoi" and "June 1953 (deep cadmium)".
\end{quote}

\begin{quote} \small
Anni Albers is American, and she  is a textile designer, a draughtsman
and a printmaker.   She  was born  in  Berlin on  12  June  1899.  She
studied art in 1916 - 1919 with Brandenburg.  Also, she studied art at
the\linebreak Kunstgewerbeschule  in Hamburg  in 1919  - 1920 and  the
Bauhaus at Weimar and Dessau  in 1922 - 1925 and  1925 - 1929. In 1933
she settled in the USA.  In 1933 -  1949 she taught at Black  Mountain
College in North Carolina.
\end{quote}}

\caption{Cumulative type use with sentences from the short
biography text type}
\label{longer-stats}
\rule{\columnwidth}{0.2mm}
\end{figure}

In the training   phase, the information   system runs with  the  full
generation grammar.  All grammatical  types used during this stage are
collected to yield the cumulative type set {\em goal-types}.  How many
text examples must be generated in this phase  depends on the relative
increase  of new  information  (occurrence of new  types) obtained with
every additional sentence generated. We show  here the results for two
related text types: `short  artist biographies' and `artist  biography
notes'. 

Figure~\ref{longer-stats} shows  the growth  curve   for the type  set
(vertical axis)   with each additional  semantic  specification passed
from the text planner to the sentence  generator (horizontal axis) for
the  first of these  text types.  The graph  shows the cumulative type
usage for   the first  90 biographies   generated, involving  some 230
sentences.\footnote{This  represented   the  current  extent   of  the
knowledge base when the test  was performed.  It is therefore possible
that with more  texts, the size  of the cumulative  set would increase
slightly since  the  curve has not  quite  `flattened  out'.  Explicit
procedures  for handling this  situation   are described below.}   The
subgrammar extraction for the ``short  artist biographies'' text  type
can therefore  be  performed with respect to   the 246 types that  are
required   by the generated   texts,  applying the algorithm described
above.  The resulting extracted subgrammar is a type lattice with only
144  types.  The size of  the  extracted subgrammar  is  only
11\% of that   of the  original  grammar.   Run  times for   sentence
generation with this extracted grammar typically range from 55\%--75\%
of that of the full grammar (see Table~\ref{runtime})---in most cases,
therefore, less   than  one second  with the  regular  KPML generation
environment   (i.e.,  unoptimized   with   full  debugging  facilities
resident).

\begin{table*}
\rule{\textwidth}{0.2mm}
\begin{center}
\begin{tabular}{l|l|l|l|p{2in}}
 & \multicolumn{2}{|c|}{run time (in ms)} & \\ 
 & full grammar & subgrammar &  improvement & sentence \\  \hline
worst case & 380 & 300 & 80 & \small ``There is Paul Delaroche.'' \\ 
best case & 3250 & 1830 & 1430 & \small ``John Foster  was  born  in  Liverpool  on  1
 January c 1787,  and  he died  at  Birkenhead  on  21 August 1846.''  \\ 
average case & ca. 900 & ca. 590 & 310 & \small \em e.g., \rm ``Mary Moser was an English
 painter.'' ``George Richmond studied  at  Royal Academy  in  1824.''
 \\ \hline
\end{tabular}

\hspace*{\fill}(Under Allegro Common Lisp running on a Sparc10.)
\end{center}
\caption{Example run times for ``short artist biographies''}
\label{runtime}
\rule{\textwidth}{0.2mm}
\end{table*}

The  generation times are  indicative    of the  style of   generation
implemented by KPML.   Clause types with  more subtypes are  likely to
cause longer  processing times than those  with  fewer subtypes.  When
there are in any case fewer subtypes available in the full grammar (as
in the existential shown in Table~\ref{runtime}), then there will be a
less noticeable  improvement compared with  the extracted grammar.  In
addition, the run  times reflect the fact that  the  number of queries
being  asked by  choosers has not  yet been  maximally  reduced in the
current  evaluation.  Noting the  cumulative set  of inquiry responses
during  the  training phase would   provide sufficient information for
more 
effective pruning of the extracted choosers.

\begin{figure}
\rule{\columnwidth}{0.2mm}

\begin{center}
\begin{picture}(210,100)
\put(0,0){\scriptsize\sffamily \grid(210,100)(30,10)[0,120]}
\thicklines
\put(0,-120){\putfile{shorts.put}{\tiny $\times$}}
\end{picture}
\end{center}

Example text:
\begin{quote} \small
Nathan Drake was an English painter. He was born at Lincoln in 1728, and
he died at York on 19 February 1778.
\end{quote}
\caption{Cumulative type use with sentences from the note
biography text type}
\label{eg2}
\rule{\columnwidth}{0.2mm}
\end{figure}

The  second   example  shows  similar improvements.   The   very short
biography entry is appropriate more for figure headings, margin notes,
etc.  The   cumulative type use graph   is  shown in Figure~\ref{eg2}.
With  this `smaller' text  type,  the  cumulative use stabilizes  very
quickly (i.e.,  after 39  sentences)  at 205  types.  This remained
stable for a test set of  500 sentences.  Extracting the corresponding
subgrammar yields a grammar involving only 101  types, which is 7\% of
the original grammar.  Sentence generation time is accordingly faster,
ranging from 40\%--60\% of that of the full grammar. In both cases, it
is clear that the size   of the resulting subgrammar is   dramatically
reduced. The generation run-time is cut to 2/3. The run-time    space  requirements are cut   similarly.  The
processing time for subgrammar extraction is less than one minute, and
is therefore not a significant issue for improvement.

\section{Conclusions and discussion}

In this paper,   we    have described how  generation   resources  for
restricted applications    can  be developed  drawing   on large-scale
general   generation grammars.  This  enables   both  re-use of  those
resources and  progressive  growth as new applications   are met.  The
grammar extraction tool then  makes it a  simple task to  extract from
the large-scale  resources specially tuned subgrammars  for particular
applications.  Our approach shows  some similarities to  that proposed
by \cite{RaynerCarter96}  for improving parsing performance by grammar
pruning and specialization  with respect to a  training  corpus.  Rule
components     are `chunked' and  pruned   when   they are unlikely to
contribute to a successful parse.  Here we have shown how improvements
in generation  performance can be achieved  for generation grammars by
removing parts of the grammar specification  that are not used in some
particular sublanguage.  The extracted  grammar is generally  known to
cover  the target  sublanguage and so  there  is  no  loss of required
coverage. 

Another motivation  for this work  is  the need  for smaller, but  not
toy-sized, systemic grammars  for their experimental compilation  into
state-of-the-art  feature logics.   The   ready access to   consistent
subgrammars of   arbitrary size given   with  the automatic subgrammar
extraction reported here allows us  to investigate further the size to
which feature logic representations of systemic grammar can grow while
remaining  practically usable.  The  compilation of  the  full grammar
NIGEL  has   so far      only    proved  possible  for    CUF     (see
\cite{Henschel95-edin}),  and the resulting   type deduction runs  too
slowly for practical applications.

It is likely that  further improvements in generation performance will
be  achieved when both the  grammatical  structures and the  extracted
choosers are pruned. The current results have focused primarily on the
improvements brought by   reconfiguring the type lattice that  defines
the grammar. The structures generated are still the `full' grammatical
structures that are produced by   the corresponding full grammar:  if,
however,   certain    constituent  descriptions   are   always unified
(conflated     in systemic        terminology)    then,    analogously
to~\cite{RaynerCarter96},  they  are candidates for  replacement  by a
single constituent  description in the extracted subgrammar. Moreover,
the extracted choosers can also be pruned directly with respect to the
sublanguage.  Currently the pruning  carried out is only that entailed
by the type lattice, It is also possible  however to maintain a record
of the classificatory inquiry responses that are used in a subgrammar:
responses that do  not occur can then  motivate further reductions  in
the choosers that are kept in the extracted grammar. Evaluation of the
improvements   in performance   that  these  strategies  bring  are in
progress.

One  possible   benefit  of  not pruning   the chooser  decision trees
completely  is to provide a fall-back  position  for when the input to
the  generation component in fact strays  outside  of that expected by
the targetted subgrammar.  Paths in  the chooser decision tree that do
not correspond to types in the subgrammar can be maintained and marked
explicitly as  `out of bounds' for  that  subgrammar.  This provides a
semantic check that the semantic inputs to the generator remain within
the  limits inherent in the extracted  subgrammar.  If it sufficiently
clear that these limits will   be adhered to, then further  extraction
will be  free of problems.  However  if the  demands of an application
change over time, then it is also possible  to use the semantic checks
to trigger  regeneration with the full  grammar: this  offers improved
average  throughput  while  maintaining complete   generation.  Noting
exceptions can  also be used to  trigger new subgrammar extractions to
adapt  to   the new applications  demands.     A number  of strategies
therefore present themselves for incorporating grammar extraction into
the application development cycle.

Although we have focused  here on run-time  improvements, it is  clear
that the grammar extraction tool has other possible uses. For example,
the  existence of   small grammars  is  one  important contribution to
providing teaching materials.  Also, the ability to extract consistent
subcomponents should make  it more straightforward to  combine grammar
fragments as required for particular needs. Further validation in both
areas  forms part of our  ongoing research.  Moreover, a significantly
larger reduction  of the type lattice  can be expected by starting not
from the cumulative set of  goal-types for the grammar reduction,  but
from  a detailed protocol  of jointly  used  types for every generated
sentence of  the training corpus.   A clustering  technique applied to
such a protocol is under development.

Finally, the proposed procedure is  not bound to systemic grammar and
can also be used  to  extract  common typed  unification  subgrammars.
Here, however,  the  gain will probably   not be as remarkable   as in
systemic grammar.   The universal principles  of, for example, an HPSG
cannot  be excised.    HPSG type  hierarchies  usually contain  mainly
general types,  so that they  will not be  affected substantially.  In
the end, the  degree of improvement achieved depends  on the extent to
which a grammar explicitly includes in its type hierarchy distinctions
that are fine enough to vary depending on text type.

\end{document}